\documentclass[letterpaper, 10 pt, conference]{ieeeconf}  

\IEEEoverridecommandlockouts

\usepackage{url}
\usepackage{makecell}
\usepackage{multirow}
\usepackage{graphicx}
\usepackage{array}
\usepackage{tabularx}
\usepackage{booktabs}
\usepackage{amssymb}
\usepackage{pifont}
\usepackage{caption}
\usepackage[table]{xcolor}

\newcommand{\iconmark}[1]{\raisebox{-0.06ex}{\scalebox{0.92}{#1}}}
\newcommand{\cmark}{\textcolor{green!60!black}{\iconmark{\ding{51}}}}  
\newcommand{\xmark}{\textcolor{red}{\iconmark{\ding{55}}}}             
\newcommand{\pmark}{\textcolor{orange}{\iconmark{\ding{115}}}}

\definecolor{rowpi}{RGB}{220,238,246}
\definecolor{rowttt}{RGB}{246,242,214}
\definecolor{rowframe}{RGB}{216,238,222}
\definecolor{rowgsg}{RGB}{244,222,226}
\definecolor{rowmemer}{RGB}{226,226,242}
\definecolor{deltapos}{RGB}{0,128,60}
\definecolor{deltaneg}{RGB}{180,40,40}
\definecolor{deltaneu}{RGB}{120,120,120}
\DeclareRobustCommand{\neutri}{\rotatebox[origin=c]{-90}{$\blacktriangle$}}

\newcommand{\vsd}[2]{#1{\tiny\textcolor{deltaneg}{$\blacktriangledown$#2}}}
\newcommand{\vsu}[2]{#1{\tiny\textcolor{deltapos}{$\blacktriangle$#2}}}
\newcommand{\vsn}[2]{#1{\tiny\textcolor{deltaneu}{\neutri#2}}}

\title{\LARGE \bf
RoboIRGBench: Benchmarking Implicit Referential Grounding in Vision-Language-Action Models
}

\author{\authorblockN{Aernaer Akelijiang\textsuperscript{1, 2},
Jiannan Li\textsuperscript{1},
Zhineng Chen\textsuperscript{2},
Jingjing Chen\textsuperscript{2},
and Bin Zhu\textsuperscript{1 \authorrefmark{2}}}
\authorblockA{\textsuperscript{1}Singapore Management University,
\textsuperscript{2}Fudan University\\
Correspondence: binzhu@smu.edu.sg \\ 
Project Webpage: \url{https://aernar.github.io/RoboIRGBench/}
\thanks{\authorrefmark{2}Corresponding author and project lead.}}
}

\begin{document}

\bstctlcite{IEEEexample:BSTcontrol}

\IEEEaftertitletext{%
\begin{minipage}{\textwidth}
\centering
\includegraphics[width=\textwidth]{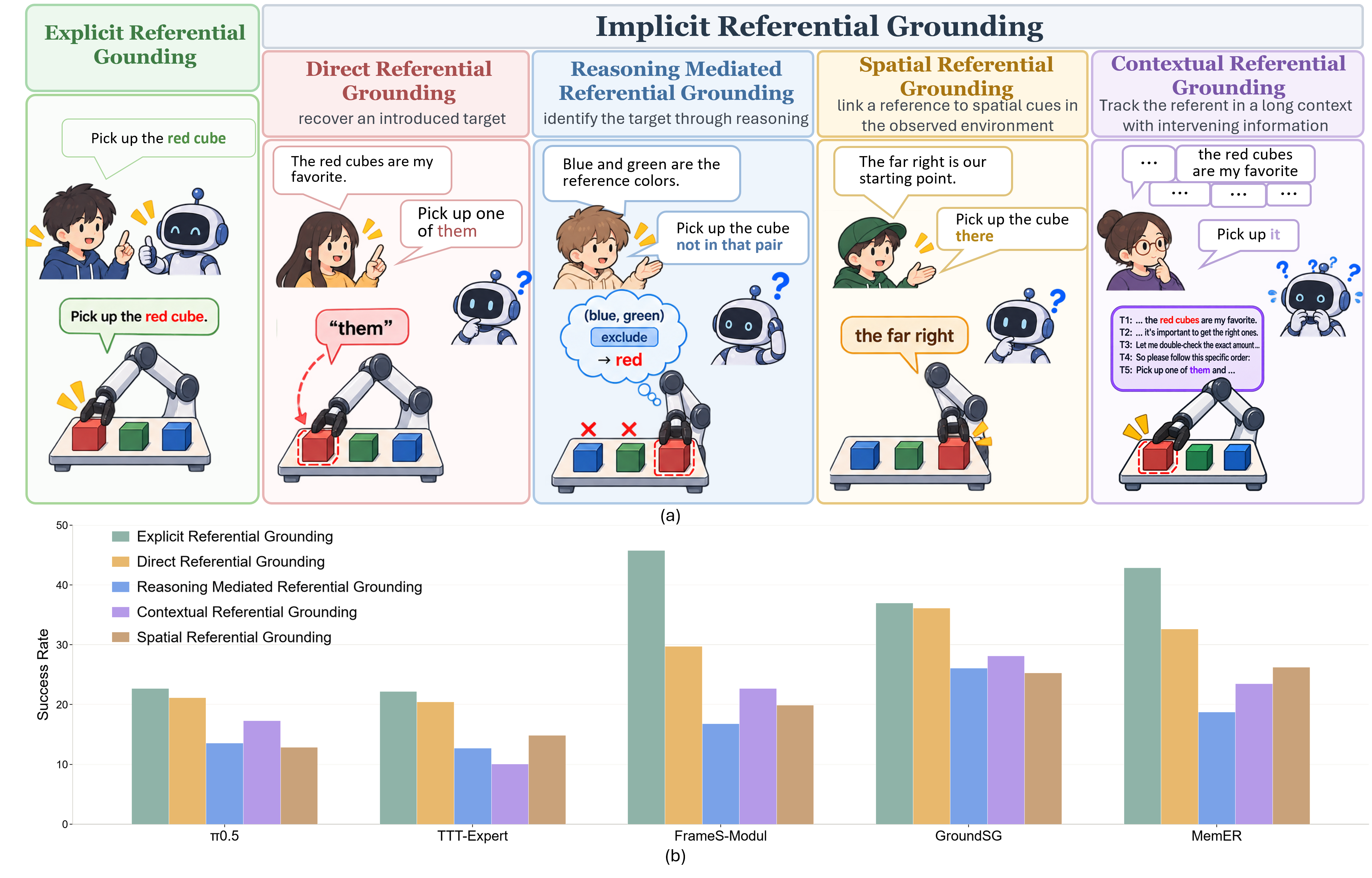}
\captionsetup{type=figure,skip=0pt}
\caption{{Overview of RoboIRG-Bench.} (a) Benchmark design covering conventional Explicit Referential Grounding and four forms of Implicit Referential Grounding (IRG): Direct, Reasoning-Mediated, Spatial, and Contextual Referential Grounding. The examples illustrate how each setting requires different forms of referent resolution from linguistic or perceptual context. (b) Average success rates for all tasks of representative memory VLA models across the explicit and four IRG settings, revealing a noticeable performance gap under implicit referential grounding.}
\label{fig:main}
\end{minipage}
}

\maketitle
\thispagestyle{empty}
\pagestyle{empty}

\begin{abstract}

Vision-Language-Action (VLA) models have shown strong capabilities in robotic manipulation, yet existing benchmarks typically assume that task-relevant information is explicitly specified in the instruction. In practice, however, humans frequently refer to objects, quantities, and relations implicitly, requiring robots to recover the intended target from linguistic and perceptual context. We study this capability as \textbf{Implicit Referential Grounding (IRG)} and introduce \textbf{RoboIRG-Bench}, a manipulation benchmark designed to systematically evaluate it. Built upon RoboMME, RoboIRG-Bench contains 40 variants derived from 11 tasks and covers four challenges, including direct, reasoning-mediated, spatial, and contextual referential grounding. As IRG often requires retaining and retrieving previously established context, we evaluate representative VLAs spanning different memory mechanisms. Our evaluation reveals a noticeable referential robustness gap. Models that perform well under explicit instructions can degrade sharply when the same task-relevant information must be recovered from context. Reasoning-mediated and spatial references are particularly challenging, while models using external VLMs show greater robustness but still exhibit significant failures. Moreover, replacing the external VLM with a stronger model does not eliminate these gaps. We further validate these findings on a Franka Research 3 robot arm, where the gap persists under real-world manipulation and manifests as both incorrect referent grounding and downstream execution failures. These results establish IRG as a distinct and underexplored capability for reliable robotic instruction following and highlight the need for VLAs that can robustly integrate language, perception, reasoning, and action.

\end{abstract}

\section{INTRODUCTION}

Vision-Language-Action (VLA) models~\cite{octo,pi05,memer} have recently shown substantial progress in robotic manipulation, demonstrating increasingly strong capabilities in long-horizon task execution, knowledge generalization, and memory-based reasoning. Alongside this progress, a growing number of benchmarks have been developed to evaluate increasingly complex manipulation capabilities~\cite{RoboCerebra,VLABench,Robocasa, robomme}. However, most existing benchmarks make a simplifying assumption that task-relevant information is explicitly stated in the instruction. In natural human–robot interaction, this assumption is often unrealistic. Humans routinely use expressions such as “it,” “the one there,” or “the other one,”, relying on linguistic and perceptual context rather than repeatedly naming every object, quantity, or relation.

We study this problem through Implicit Referential Grounding (IRG): the ability of a robot to identify task-relevant entities, attributes, quantities, or relations when they are not explicitly specified in the current instruction and must instead be resolved from linguistic or perceptual context. For example, after hearing “the red cubes are my favorite,” a robot receiving “put one of them into the bin” must determine that “them” refers to the red cubes before executing the manipulation.  More challenging cases may require inferring a target through exclusion or numerical relations, grounding a reference through spatial cues, or maintaining the relevant information across extended contextual descriptions. IRG is closely related to memory~\cite{robomme, memer}, but the two are not equivalent. Resolving an implicit reference often requires retaining or retrieving previously introduced information, making memory an important prerequisite. However, memory alone is insufficient. The robot must determine which contextual information the current expression refers to, reason over it when necessary, ground it to the observed scene, and translate the resolved referent into action. Thus, memory determines what contextual information is available, while IRG determines how that information is interpreted and grounded for control. This capability is important because successful instruction following under explicit language or memory mechanisms does not necessarily imply robust referential grounding. A model may reliably execute “pick up the red cube” yet fail when the same target must be inferred from “pick up the one I mentioned earlier.” Such failures are largely underexplored by manipulation and memory benchmarks~\cite{RLBench,libero,RoboTwin2.0, robomme, RoboMemArena}, where task-relevant entities are typically specified explicitly, as listed in Table~\ref{tab:re}.

To address this gap, we introduce RoboIRGBench, a benchmark for systematically evaluating implicit referential grounding for VLAs in robotic manipulation. Built upon the RoboMME~\cite{robomme}, RoboIRG-Bench contains 40 task variants derived from 11 manipulation tasks and spans three difficulty levels. As illustrated in Figure \ref{fig:main}, we organize the benchmark into four complementary challenges. \textit{Direct Referential Grounding} requires recovering a previously introduced target. \textit{Reasoning-Mediated Referential Grounding} requires identifying the target through operations such as color exclusion, numerical deduction, ordinal relations, or contextual logic. \textit{Spatial Referential Grounding} requires linking a reference to spatial cues in the observed environment. \textit{Contextual Referential Grounding} evaluates whether the model can retain and recover a referent despite intervening linguistic context. For each setting, we modify how task-defining information is expressed while preserving the underlying manipulation objective wherever possible. The affected variables include object properties, repetition counts, temporal relations, and spatial relations, while the corresponding action descriptions remain unchanged.

Our evaluation reveals a noticeable gap between explicit instruction-following capability and implicit referential robustness across representative VLA architectures with different memory mechanisms. Models that perform strongly under conventional instructions can degrade sharply when the task-relevant information is expressed implicitly, showing that memory alone does not guarantee robust referential grounding. Models incorporating an external Vision-Language Model (VLM) (e.g., Qwen-VL) for subgoal generation show relatively stronger performance, but substantial failures remain when reference resolution requires semantic reasoning, spatial understanding, or extended-context tracking. Moreover, replacing the external VLM with a substantially stronger model (e.g., Gemini 3.1 Pro) does not eliminate these failures, indicating that simply scaling the upstream vision-language component is insufficient to guarantee robust referential grounding. We further validate these findings in real-world manipulation on a Franka Research 3 robot arm, where the same robustness gap persists under physical execution. The observed failures include both incorrect referent grounding and downstream execution errors, showing that IRG remains challenging beyond simulation.

Our main contributions are summarized as follows:
\begin{itemize}
    \item We formulate implicit referential grounding as an important yet underexplored evaluation dimension for VLA models, examining whether robotic policies can recover task-relevant information from linguistic and perceptual context rather than relying on explicit target specification.

    \item We introduce RoboIRG-Bench, a simulation manipulation benchmark comprising 40 task variants across four complementary challenges: direct, reasoning-mediated, spatial, and contextual referential grounding.

    \item We conduct a systematic paired evaluation of explicit and implicit instructions, revealing substantial referential robustness gaps across representative VLA systems and showing that stronger external vision-language reasoning alone does not eliminate these failures. We further validate this robustness gap through real-world manipulation experiments on a Franka robot.

\end{itemize}

\section{Related Works}

\subsection{Robotic Manipulation Benchmarks}

\begin{table}
\caption{
Comparison of Implicit Referential Grounding (IRG) coverage across existing
robotic manipulation benchmarks.
\cmark~covered; \xmark~not covered;
\pmark~partially covered.
}
\label{tab:re}
\centering
\scriptsize
\setlength{\tabcolsep}{1.6pt}
\renewcommand{\arraystretch}{0.98}

\begin{tabular}{@{}lcccc@{}}
\hline\hline
Dataset &
\makecell{Direct} &
\makecell{Reasoning-Mediated} &
\makecell{Spatial} &
\makecell{Contextual} \\
\hline

RLBench~\cite{RLBench}
    & \pmark & \xmark & \xmark & \xmark \\
CALVIN~\cite{CALVIN}
    & \pmark & \xmark & \xmark & \xmark \\
ARNOLD~\cite{ARNOLD}
    & \xmark & \xmark & \xmark & \xmark \\
LIBERO~\cite{libero}
    & \pmark & \xmark & \xmark & \xmark \\
RoboCasa~\cite{Robocasa}
    & \pmark & \xmark & \xmark & \xmark \\
VLABench~\cite{VLABench}
    & \cmark & \xmark & \xmark & \xmark \\
RoboTwin 2.0~\cite{RoboTwin2.0}
    & \pmark & \xmark & \xmark & \xmark \\
RoboCerebra~\cite{RoboCerebra}
    & \cmark & \xmark & \xmark & \xmark \\

\hline

MemoryBench~\cite{sam2act}
    & \xmark & \xmark & \xmark & \xmark \\
MIKASA-robo (VLA)~\cite{mikasa}
    & \pmark & \xmark & \xmark & \xmark \\
RoboMME~\cite{robomme}
    & \pmark & \xmark & \xmark & \xmark \\
RMBench~\cite{RMBench}
    & \cmark & \xmark & \xmark & \xmark \\
RoboMemArena~\cite{RoboMemArena}
    & \xmark & \xmark & \xmark & \xmark \\
\textbf{RoboIRGBench}
    & \cmark & \cmark & \cmark & \cmark \\

\hline\hline
\end{tabular}
\end{table}

Recently, various robotic simulation benchmarks~\cite{Robocasa,RoboCerebra,VLABench, li2026libero} have advanced robotic manipulation evaluation in scene diversity, long-horizon planning, knowledge generalization and robustness. More recently, memory-oriented benchmarks (below the line in Table \ref{tab:re}) have emerged to incorporate memory capabilities into robotic manipulation evaluation.

However, as summarized in Table~\ref{tab:re}, existing manipulation benchmarks generally specify task-relevant entities explicitly, rather than evaluating how contextual information is interpreted and grounded for control. Although some benchmarks include pronouns (marked by \pmark), these pronouns mainly appear in sequential action chains such as ``pick up X, then place it,'' rather than requiring the robot to resolve implicit references from linguistic or perceptual context.

\subsection{Implicit Referential Grounding}

Implicit referential expressions are used extensively in human dialogue. 
Cognitive studies indicate that reduced referring expressions, such as pronouns, occur more frequently when the referent is mentioned more often in prior discourse or appears in recent linguistic context~\cite{Ariel_cog, Givon_cog}. 
The ability to resolve these expressions—which we term \textit{implicit referential grounding}—is a fundamental human capability. 
Recent benchmarks, such as REI-Bench~\cite{rei-bench}, have begun addressing implicit referential grounding within robot task planning scenarios. However, their evaluations rely on action primitives and scaled-up textual complexity, failing to capture diverse referential scenarios present in direct robotic manipulation.

\section{RoboIRGBench Construction}

\subsection{Benchmark Overview} 

We introduce RoboIRGBench, a benchmark for evaluating Implicit Referential Grounding (IRG) in closed-loop robotic manipulation. RoboIRG-Bench is built upon RoboMME~\cite{robomme} and contains 40 task variants derived from 11 manipulation tasks from the Counting, Permanence, and Reference suites of RoboMME. These tasks are chosen because their goals contain explicit task-defining variables that can be systematically reformulated as implicit references, including object properties, repetition counts, temporal relations and spatial relations. We exclude five RoboMME tasks whose instructions do not contain such variables. In these tasks, the manipulation objective is determined primarily by the visual scene itself, making it difficult to construct an implicit-reference counterpart without changing the task semantics. Each task variant is evaluated at three difficulty levels (Easy, Medium, and Hard) using rule-based instruction templates to ensure controlled and reproducible language generation.

The central design principle is to preserve the underlying manipulation objective while changing how task-relevant information is specified. Instead of directly naming the target entity, quantity, or relation, RoboIRGBench requires the robot to recover it from preceding linguistic context or the observed scene. Since such recovery often depends on previously established context, the benchmark naturally probes the interface between memory and referential grounding. Memory determines what contextual information is retained or retrieved, while IRG evaluates whether that information can be correctly interpreted and grounded for action. We organize these challenges into four complementary dimensions: \textit{Direct Referential Grounding, Reasoning-Mediated Referential Grounding, Spatial Referential Grounding, and Contextual Referential Grounding.}

\subsection{Implicit Referential Grounding Challenges}

\begin{figure}
\centering
\includegraphics[width=\columnwidth]{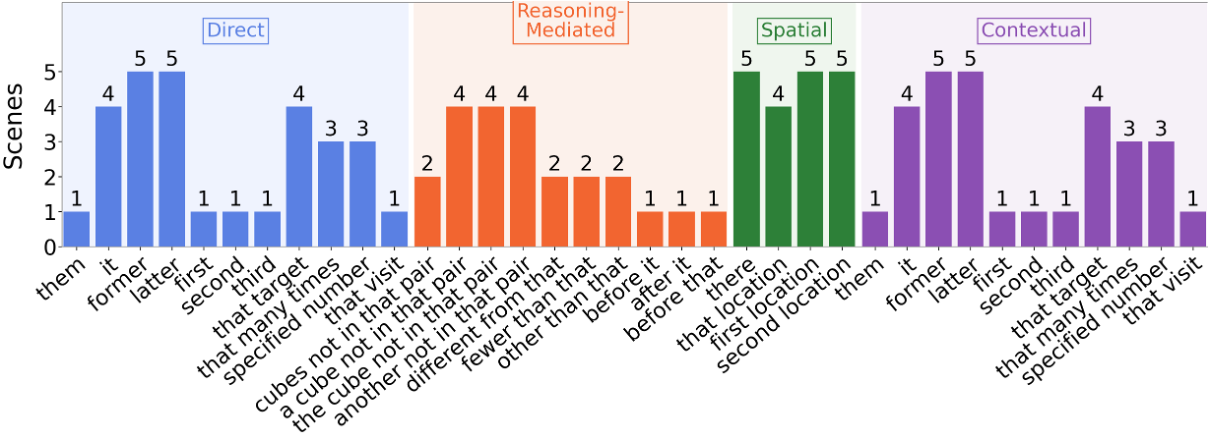}
\caption{Distribution of referring expressions over scenes, grouped by the four implicit referential grounding strategies. Each expression is counted once per scene.}
\label{fig:stats}
\end{figure}

\textbf{Direct Referential Grounding.} This setting evaluates the most basic form of referential grounding, which requires recovering a target that has been explicitly introduced in preceding context but is not repeated in the final instruction. For example, after introducing “the red cubes are my favorite,” the subsequent command may ask the robot to “put one of them into the bin.” We replace explicit target mentions with reduced referring expressions, including pronouns such as it and them, as well as nominal references such as the former, the latter, the first/second/third, and that. Figure~\ref{fig:stats} summarizes the distribution of these referring expressions across scenes under the four implicit referential grounding strategies. Successful execution therefore requires the model to retain the previously introduced target and correctly bind the final reference to it.

\textbf{Reasoning-Mediated Referential Grounding.} Direct referential grounding is insufficient for many natural instructions. The intended target may need to be inferred from a referenced entity, quantity, or event. We therefore construct a more challenging setting in which the preceding context specifies an intermediate reference, while the final target must be derived through additional reasoning. We instantiate four complementary reasoning patterns. \textit{(1) Color Exclusion.} The context specifies one or more reference colors, while the robot must identify an object whose color satisfies an exclusion relation, such as “pick up a cube whose color is not in that pair.” Resolving the target requires jointly interpreting the linguistic constraint and observing the colors present in the scene. \textit{(2) Number Deduction.} The context provides a reference quantity, while the final instruction specifies the required action count relationally, e.g., “repeat the action one time fewer than that.” The model must recover the referenced quantity and perform a numerical transformation. \textit{(3) Sequential Relations.} The target is defined relative to another element in an ordered sequence, requiring the model to recover temporal or ordinal relationships such as before and after. \textit{(4) Contextual Logic.} The target must be inferred through a complementary or relational condition established by the preceding context, rather than being directly recoverable from a single mention, such as "...other than that".

\textbf{Spatial Referential Grounding.} Implicit references can also depend on the physical environment rather than linguistic context alone. We therefore evaluate whether VLA models can ground expressions such as “the cube there” when the referenced location has previously been specified through a spatial landmark.  Spatial landmarks are determined from object positions in the robot coordinate frame. In Swap tasks, landmarks are defined according to the objects' initial positions before manipulation begins. We apply this setting to seven of the eleven selected tasks in RoboMME. StopCube, VideoRepick, VideoPlaceButton, and VideoPlaceOrder are excluded because their targets are defined primarily by action counts or temporal information rather than spatial cues.

\textbf{Contextual Referential Grounding.} In real human-robot interaction, the relevant reference may be separated from the final command by additional dialogue or contextual information. We therefore introduce Contextual Referential Grounding to evaluate whether VLA models can retain and recover the correct referent under extended and potentially distracting linguistic context. Starting from the Direct Referential Grounding setting, we construct a five-stage context sequence. Starting from Direct Referential Grounding, we progressively insert contextual information between the initial referent and the final instruction. Each example follows a five-stage structure, including \textit{(1) Referent Introduction} establishing the task-relevant entity or information, (2) \textit{Referent Reinforcement} providing an additional description of the same referent, \textit{(3) Distractor Context} introducing task-irrelevant information that shifts attention away from the referent, \textit{(4) Action Cue} indicating that the manipulation instruction is about to follow. and \textit{(5) Referential Instruction} giving the final command using an implicit reference instead of explicitly restating the target.

\section{Experiment}

\begin{table*}[t]
        \caption{Success rates (\%) of representative VLA models under explicit instructions and four Implicit Referential Grounding (IRG) challenges.
        \textcolor{deltaneg}{$\blacktriangledown$}/\textcolor{deltapos}{$\blacktriangle$}: decrease/increase relative to Explicit.
        “–” indicates not applicable.}
        \label{tab:main}
        \begin{center}
        \footnotesize
        \setlength{\tabcolsep}{1.2pt}
        \setlength{\extrarowheight}{3pt}
        \resizebox{\textwidth}{!}{%
        \begin{tabular}{@{}lccccccccccccc@{}}
        \hline\hline
         & & \multicolumn{4}{c}{Counting} & \multicolumn{4}{c}{Permanence} & \multicolumn{3}{c}{Reference} & \multirow{3}{*}{Avg} \\
        \cmidrule(lr){3-6}\cmidrule(lr){7-10}\cmidrule(lr){11-13}
         & & \begin{tabular}[c]{@{}c@{}}Bin\\Fill\end{tabular}
         & \begin{tabular}[c]{@{}c@{}}Pick\\Xtimes\end{tabular}
         & \begin{tabular}[c]{@{}c@{}}Swing\\Xtimes\end{tabular}
         & \begin{tabular}[c]{@{}c@{}}Stop\\Cube\end{tabular}
         & \begin{tabular}[c]{@{}c@{}}Video\\Unmask\end{tabular}
         & \begin{tabular}[c]{@{}c@{}}Button\\Unmask\end{tabular}
         & \begin{tabular}[c]{@{}c@{}}Video\\UnmaskS\end{tabular}
         & \begin{tabular}[c]{@{}c@{}}Button\\UnmaskS\end{tabular}
         & \begin{tabular}[c]{@{}c@{}}Video\\Repick\end{tabular}
         & \begin{tabular}[c]{@{}c@{}}Video\\PlaceButton\end{tabular}
         & \begin{tabular}[c]{@{}c@{}}Video\\PlaceOrder\end{tabular} \\
        \hline
        \rowcolor{rowpi} & $\pi_{0.5}$ &28.67  &44.67  &35.33  &6.00  &23.33  &28.00  &20.67  &7.33  &0.00  &30.67  &25.33  &22.73  \\
        \rowcolor{rowttt} & TTT-Expert &27.33  &30.00  &30.67  &6.67  &34.67  &22.00  &20.67  &14.67  &7.33  &26.67  &24.00  &22.24  \\
        \rowcolor{rowframe} & FrameS-Modul &40.67  &\textbf{90.00}  &\textbf{93.33}  &\textbf{50.00}  &33.33  &28.00  &24.00  &15.33  &\textbf{30.67}  &\textbf{56.67}  &\textbf{42.00}  &\textbf{45.82}  \\
        \rowcolor{rowgsg} & GroundSG-QwenVL &\textbf{55.33}  &88.00  &4.00  &0.00  &\textbf{90.67}  &23.33  &26.00  &15.33  &29.33  &42.67  &32.67  &37.03  \\
        \rowcolor{rowmemer} \multirow{-5}{*}{\rotatebox{90}{Explicit}} & MemER &52.67  &71.33  &56.00  &2.00  &81.33  &\textbf{74.00}  &\textbf{34.00}  &\textbf{18.00}  &24.67  &26.00  &32.00  &42.91  \\
        \hline
        \rowcolor{rowpi} & $\pi_{0.5}$ & \vsd{23.33}{5.34} & \vsd{29.33}{15.34} & \vsu{37.33}{2.00} & 6.00 & \vsd{21.33}{2.00} & \vsd{26.67}{1.33} & \vsu{22.00}{1.33} & \vsu{10.67}{3.34} & \vsu{1.33}{1.33} & \vsu{34.00}{3.33} & \vsd{20.67}{4.66} & \vsd{21.15}{1.58} \\
        \rowcolor{rowttt} & TTT-Expert & \vsd{26.00}{1.33} & \vsd{19.33}{10.67} & \vsu{32.00}{1.33} & \vsd{4.67}{2.00} & \vsd{33.33}{1.34} & \vsd{18.00}{4.00} & \vsd{18.67}{2.00} & \vsu{\textbf{16.67}}{2.00} & \vsd{5.33}{2.00} & \vsd{25.33}{1.34} & \vsu{26.00}{2.00} & \vsd{20.48}{1.76} \\
        \rowcolor{rowframe} & FrameS-Modul & \vsd{28.67}{12.00} & \vsd{12.67}{77.33} & \vsd{\textbf{58.67}}{34.66} & \vsd{\textbf{31.33}}{18.67} & \vsd{29.33}{4.00} & \vsd{24.67}{3.33} & \vsu{24.67}{0.67} & \vsd{14.00}{1.33} & \vsd{10.00}{20.67} & \vsd{\textbf{50.67}}{6.00} & \vsu{\textbf{42.67}}{0.67} & \vsd{29.76}{16.06} \\
        \rowcolor{rowgsg} & GroundSG-QwenVL & \textbf{55.33} & \vsd{\textbf{85.33}}{2.67} & \vsu{8.00}{4.00} & 0.00 & \vsd{\textbf{89.33}}{1.34} & \vsd{22.00}{1.33} & \vsu{\textbf{35.33}}{9.33} & \vsd{11.33}{4.00} & \vsd{\textbf{24.67}}{4.66} & \vsd{41.33}{1.34} & \vsd{25.33}{7.34} & \vsd{\textbf{36.18}}{0.85} \\
        \rowcolor{rowmemer} \multirow{-5}{*}{\rotatebox{90}{Direct}} & MemER & \vsd{46.67}{6.00} & \vsd{44.67}{26.66} & \vsd{20.67}{35.33} & \vsd{0.00}{2.00} & \vsu{85.33}{4.00} & \vsd{\textbf{64.00}}{10.00} & \vsd{26.67}{7.33} & \vsd{4.00}{14.00} & \vsd{12.67}{12.00} & \vsd{19.33}{6.67} & \vsu{35.33}{3.33} & \vsd{32.67}{10.24} \\
        \hline
        \rowcolor{rowpi} & $\pi_{0.5}$ & \vsd{4.00}{24.67} & \vsd{0.00}{44.67} & \vsd{0.00}{35.33} & \vsd{0.67}{5.33} & \vsu{28.00}{4.67} & \vsd{24.67}{3.33} & \vsu{22.67}{2.00} & \vsu{9.33}{2.00} & \vsu{1.33}{1.33} & \vsu{31.33}{0.66} & \vsu{27.33}{2.00} & \vsd{13.58}{9.15} \\
        \rowcolor{rowttt} & TTT-Expert & \vsd{6.00}{21.33} & \vsd{0.00}{30.00} & \vsd{0.00}{30.67} & \vsd{4.00}{2.67} & \vsd{24.00}{10.67} & 22.00 & \vsd{19.33}{1.34} & \vsd{14.00}{0.67} & \vsd{3.33}{4.00} & \vsu{27.33}{0.66} & \vsu{26.67}{2.67} & \vsd{13.33}{8.91} \\
        \rowcolor{rowframe} & FrameS-Modul & \vsd{10.67}{30.00} & \vsd{0.00}{90.00} & \vsd{0.00}{93.33} & \vsd{\textbf{12.67}}{37.33} & \vsd{23.33}{10.00} & \vsd{\textbf{26.00}}{2.00} & \vsd{19.33}{4.67} & \vsd{\textbf{14.67}}{0.66} & \vsd{14.00}{16.67} & \vsd{28.00}{28.67} & \vsd{36.67}{5.33} & \vsd{16.85}{28.97} \\
        \rowcolor{rowgsg} & GroundSG-QwenVL & \vsd{\textbf{46.67}}{8.66} & \vsd{\textbf{66.00}}{22.00} & 4.00 & 0.00 & \vsd{\textbf{48.00}}{42.67} & \vsd{19.33}{4.00} & \vsd{16.00}{10.00} & \vsd{10.00}{5.33} & \vsd{10.67}{18.66} & \vsd{\textbf{37.33}}{5.34} & \vsd{29.33}{3.34} & \vsd{\textbf{26.12}}{10.91} \\
        \rowcolor{rowmemer} \multirow{-5}{*}{\rotatebox{90}{Reasoning-Me}} & MemER & \vsd{13.33}{39.34} & \vsd{21.33}{50.00} & \vsd{\textbf{8.67}}{47.33} & \vsd{0.00}{2.00} & \vsd{30.00}{51.33} & \vsd{22.67}{51.33} & \vsd{\textbf{23.33}}{10.67} & \vsd{6.67}{11.33} & \vsd{\textbf{16.00}}{8.67} & \vsd{23.33}{2.67} & \vsu{\textbf{41.33}}{9.33} & \vsd{18.79}{24.12} \\
        \hline
        \rowcolor{rowpi} & $\pi_{0.5}$ & \vsd{20.67}{8.00} & \vsd{28.00}{16.67} & \vsd{26.00}{9.33} & \vsd{2.67}{3.33} & \vsu{25.33}{2.00} & \vsd{12.67}{15.33} & \vsd{18.00}{2.67} & \vsd{2.00}{5.33} & \vsu{1.33}{1.33} & \vsu{\textbf{34.00}}{3.33} & \vsd{20.00}{5.33} & \vsd{17.33}{5.40} \\
        \rowcolor{rowttt} & TTT-Expert & \vsd{4.00}{23.33} & \vsd{8.00}{22.00} & \vsd{12.00}{18.67} & \vsd{2.00}{4.67} & \vsd{20.00}{14.67} & \vsd{1.33}{20.67} & \vsd{14.00}{6.67} & \vsd{0.00}{14.67} & \vsd{0.00}{7.33} & \vsd{24.67}{2.00} & \vsu{24.67}{0.67} & \vsd{10.06}{12.18} \\
        \rowcolor{rowframe} & FrameS-Modul & \vsd{21.33}{19.34} & \vsd{13.33}{76.67} & \vsd{\textbf{38.67}}{54.66} & \vsd{\textbf{7.33}}{42.67} & \vsu{36.00}{2.67} & \vsu{28.67}{0.67} & \vsu{25.33}{1.33} & \vsu{\textbf{16.67}}{1.34} & \vsd{8.00}{22.67} & \vsd{27.33}{29.34} & \vsd{27.33}{14.67} & \vsd{22.73}{23.09} \\
        \rowcolor{rowgsg} & GroundSG-QwenVL & \vsd{\textbf{43.33}}{12.00} & \vsd{\textbf{58.67}}{29.33} & \vsd{3.33}{0.67} & \vsu{3.33}{3.33} & \vsd{\textbf{65.33}}{25.34} & \vsu{26.67}{3.34} & \vsu{\textbf{28.00}}{2.00} & \vsd{10.67}{4.66} & \vsd{\textbf{22.67}}{6.66} & \vsd{22.00}{20.67} & \vsd{26.00}{6.67} & \vsd{\textbf{28.18}}{8.85} \\
        \rowcolor{rowmemer} \multirow{-5}{*}{\rotatebox{90}{Contextual}} & MemER & \vsd{34.67}{18.00} & \vsd{37.33}{34.00} & \vsd{5.33}{50.67} & \vsd{0.67}{1.33} & \vsd{56.00}{25.33} & \vsd{\textbf{42.67}}{31.33} & \vsd{24.00}{10.00} & \vsd{10.67}{7.33} & \vsd{6.67}{18.00} & \vsd{11.33}{14.67} & \vsd{\textbf{29.33}}{2.67} & \vsd{23.52}{19.39} \\
        \hline
        \rowcolor{rowpi} & $\pi_{0.5}$ & \vsd{6.00}{22.67} & \vsd{4.67}{40.00} & \vsd{12.00}{23.33} & -- & \vsd{18.67}{4.66} & \vsd{20.67}{7.33} & \vsu{21.33}{0.66} & \vsd{3.33}{4.00} & -- & -- & -- & \vsd{12.38}{10.35} \\
        \rowcolor{rowttt} & TTT-Expert & \vsd{12.00}{15.33} & \vsd{2.67}{27.33} & \vsd{6.00}{24.67} & -- & \vsd{26.67}{8.00} & \vsu{25.33}{3.33} & 20.67 & \vsd{10.67}{4.00} & -- & -- & -- & \vsd{14.86}{7.38} \\
        \rowcolor{rowframe} & FrameS-Modul & \vsd{18.67}{22.00} & \vsd{14.00}{76.00} & \vsd{\textbf{19.33}}{74.00} & -- & \vsd{24.67}{8.66} & \vsd{21.33}{6.67} & \vsd{22.00}{2.00} & \vsu{\textbf{19.33}}{4.00} & -- & -- & -- & \vsd{19.90}{25.92} \\
        \rowcolor{rowgsg} & GroundSG-QwenVL & \vsd{17.33}{38.00} & \vsd{\textbf{29.33}}{58.67} & \vsd{0.00}{4.00} & -- & \vsd{59.33}{31.34} & \vsu{\textbf{32.67}}{9.34} & \vsd{\textbf{24.67}}{1.33} & \vsd{14.00}{1.33} & -- & -- & -- & \vsd{25.33}{11.70} \\
        \rowcolor{rowmemer} \multirow{-5}{*}{\rotatebox{90}{Spatial}} & MemER & \vsd{\textbf{34.00}}{18.67} & \vsd{26.67}{44.66} & \vsd{2.00}{54.00} & -- & \vsd{\textbf{61.33}}{20.00} & \vsd{24.67}{49.33} & \vsd{\textbf{24.67}}{9.33} & \vsd{10.67}{7.33} & -- & -- & -- & \vsd{\textbf{26.29}}{16.62} \\
        \hline\hline
        \end{tabular}%
        }
        \end{center}
        \end{table*}

\begin{figure*}[t]
\centering
\includegraphics[width=\textwidth]{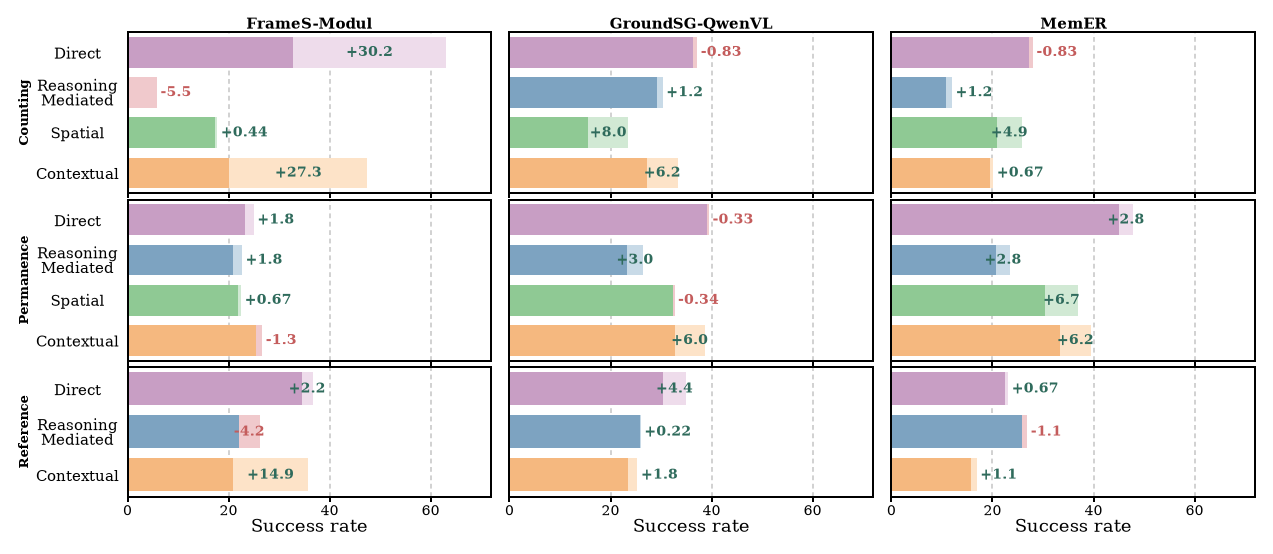}
\caption{Suite-averaged success rate. Four referential strategies with implicit expressions versus the same strategy after those expressions are made explicit. Dark shows the overlap, and $|\Delta|$ is light (explicit gain) or pink (explicit drop). $\Delta$ is annotated.}
\label{fig:explicit-ablation}
\end{figure*}

\subsection{Experimental Setup}

\textbf{Models.} IRG often requires retaining or retrieving previously established context, thus VLAs with different memory mechanisms provide a natural testbed for studying the relationship between memory and referential grounding. Following RoboMME~\cite{robomme}, we evaluate representative configurations spanning symbolic, perceptual, and recurrent memory, including GroundSG-QwenVL (\textit{symbolic + groundedSubgoal + QwenVL}), FrameS-Modul (\textit{perceptual + framesample + modul}), TTT-Expert (\textit{recurrent + ttt + expert}). All of these variants are built upon $\pi_{0.5}$. In addition, we evaluate MemER~\cite{memer}, a state-of-the-art retrieval-based memory VLA, together with vanilla $\pi_{0.5}$ as a general-purpose baseline without an explicit memory module.

\textbf{Evaluation Protocol.} For each task configuration, we conduct 50 episodes under each of three random seeds, resulting in 150 episodes per configuration. Performance is measured by task success rate (SR), averaged across the three seeds.

\subsection{Main Results and Analysis}
Table \ref{tab:main} summarizes performance across explicit and the four IRG challenges. The results reveal a clear discrepancy between conventional explicit instruction-following ability and implicit referential robustness.
First, strong performance under explicit instructions does not reliably transfer to implicit formulations. For example, FrameS-Modul achieves the highest average SR under the explicit setting (45.82\%), but drops substantially to 29.76\% under Direct Referential Grounding and 16.85\% under Reasoning-Mediated Referential Grounding. In contrast, GroundSG-QwenVL achieves the strongest overall performance on three of the four IRG challenges. MemER performs best under Spatial Referential Grounding with 26.29\%, narrowly exceeding GroundSG-QwenVL at 25.33\%. Second, Reasoning-Mediated Referential Grounding is consistently the most challenging setting. Compared with the explicit setting, average success rates decrease substantially across all evaluated models, indicating that recovering a reference becomes particularly difficult when the referent cannot be retrieved directly but must instead be derived through color exclusion, numerical deduction, sequential relations, or contextual logic.
Third, the results reveal an important distinction between memory availability and referential robustness. All memory-augmented architectures exhibit substantial degradation in many IRG settings, indicating that retaining historical information alone does not guarantee that it can be correctly referenced and grounded. GroundSG-QwenVL and MemER, both of which use an external VLM to construct subgoals, generally retain more performance under implicit references than the other evaluated systems. This observation suggests that explicit intermediate language reasoning helps implicit referential grounding.

\begin{figure*}[t]
\centering
\includegraphics[width=0.95\textwidth]{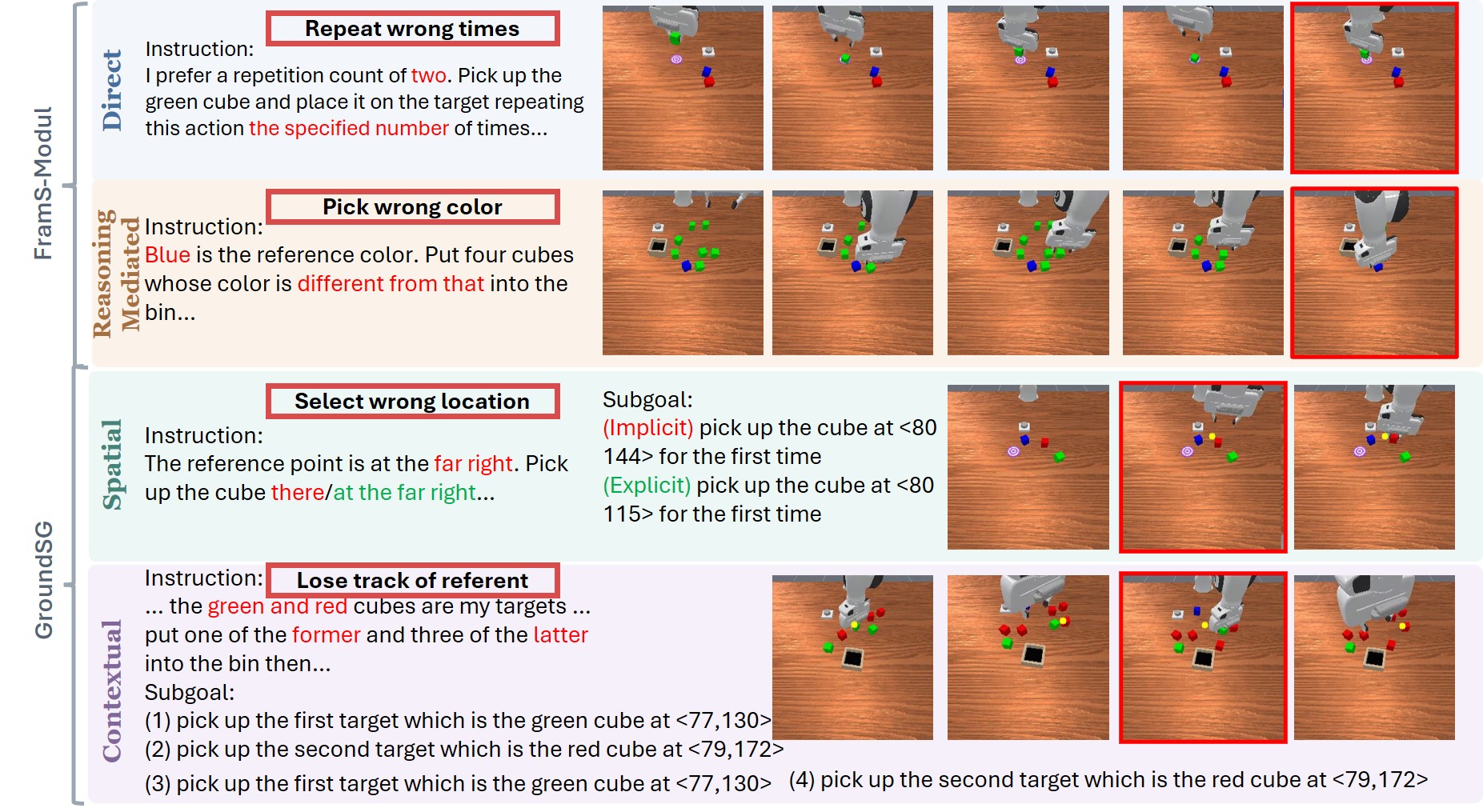}
\caption{Representative failure modes under Implicit Referential Grounding (IRG). FramS-Modul fails to resolve referenced repetition counts under Direct Referential Grounding and target identities under Reasoning-Mediated Referential Grounding. GroundSG exhibits incorrect spatial grounding and loses track of previously introduced referents under extended context, leading to erroneous subgoal generation. Red boxes highlight failures under implicit instructions.}
\label{fig:failure}
\end{figure*}

Figure \ref{fig:failure} reveals several recurring failure patterns under IRG. FramS-Modul often fails to correctly resolve or transform referential information, such as grounding the required repetition count from a numerical reference or identifying the intended target through color exclusion. For GroundSG, which uses a VLM to generate intermediate subgoals, implicit references can directly affect subgoal generation: spatial references may lead to different grounded locations, while long contexts can cause the VLM to lose track of previously established referents and generate incorrect subsequent subgoals. Overall, these failures mainly stem from weaknesses in referential resolution, implicit reasoning, spatial grounding, and long-context referent tracking.

\subsection{Isolating the Effect of Implicit Referential Grounding}
Absolute task success alone cannot determine whether a failure arises from reference resolution or from the underlying perception and control problem. We therefore construct a paired explicit counterpart for each implicit instruction while preserving the surrounding context and manipulation objective. For instance, “the cube there” is replaced by “the cube at the far right,” “different from that” by “different from blue,” while in “one time fewer than that”, “that” is replaced by the corresponding explicit quantity. Figure~\ref{fig:explicit-ablation} reports the success rate comparison between explicit instructions and their implicit counterparts at the suite level. Making the target explicit substantially improves performance in many tasks and settings, confirming that a meaningful portion of the observed degradation originates from referential grounding rather than manipulation difficulty alone. The effect is particularly obvious for FrameS-Modul under Direct Referential Grounding and Contextual Referential Grounding. Spatial Referential Grounding likewise benefits from explicit target specification, indicating that resolving a referenced spatial landmark introduces an additional challenge beyond executing the corresponding manipulation. The results under Reasoning-Mediated Referential Grounding are more heterogeneous. Explicitly resolving the reference does not always recover performance and can occasionally reduce task success. This indicates that these tasks combine referential grounding with additional semantic or numerical reasoning, and that eliminating the reference-resolution component alone does not necessarily remove the remaining reasoning bottleneck.

\subsection{Does a Stronger External VLM Solve IRG?}

The relatively strong performance of GroundSG-QwenVL raises a natural question: can IRG failures be mitigated simply by using a more capable external VLM? To investigate this, we replace the external VLM in GroundSG with Gemini 3.1 Pro and evaluate three representative manipulation tasks, including BinFill, PickXTimes, and SwingXTimes. As shown in Table \ref{tab:gemini-qwenvl}, compared with GroundSG-QwenVL, GroundSG-Gemini achieves notably higher average success rates under Reasoning-Mediated (54.44\% vs. 38.89\%), Contextual (51.34\% vs. 35.11\%), and Spatial Referential Grounding (42.22\% vs. 15.55\%). However, the referential robustness gap persists even with Gemini 3.1 Pro. Relative to its own Explicit performance (55.33\%), Gemini still drops by 6.44\%, 0.89\%, 3.99\%, and 13.11\% under Direct, Reasoning-Mediated, Contextual, and Spatial Referential Grounding, respectively. Spatial grounding remains particularly challenging despite the stronger VLM. These results show that scaling the external VLM mitigates but does not eliminate IRG failures, suggesting that robust IRG requires tighter integration of language, vision, reasoning, and action.

\begin{table}
        \caption{Performance Comparison between QwenVL and Gemini 3.1 Pro across three representative tasks under Explicit and four IRG settings in terms of success rate.
        \textcolor{deltaneg}{$\blacktriangledown$}/\textcolor{deltapos}{$\blacktriangle$}/\textcolor{deltaneu}{\neutri}:
        decrease/increase/no change relative to Explicit.}
        \label{tab:gemini-qwenvl}
        \begin{center}
        \scriptsize
        \setlength{\tabcolsep}{2.2pt}
        \setlength{\extrarowheight}{2pt}
        \resizebox{\columnwidth}{!}{%
        \begin{tabular}{@{}llcccc@{}}
        \hline\hline
        Model & Strategy & BinFill & PickXtimes & SwingXtimes & Avg \\
        \hline
        \multirow{5}{*}{QwenVL}
         & Explicit   & 55.33 & 88.00 & 4.00 & 49.11 \\
         \cline{2-6}
         & Direct     & \vsn{55.33}{0.00} & \vsd{85.33}{2.67} & \vsu{8.00}{4.00} & \vsu{49.55}{0.44} \\
         & Reasoning-Me  & \vsd{46.67}{8.66} & \vsd{66.00}{22.00} & \vsn{4.00}{0.00} & \vsd{38.89}{10.22} \\
         & Contextual & \vsd{43.33}{12.00} & \vsd{58.67}{29.33} & \vsd{3.33}{0.67} & \vsd{35.11}{14.00} \\
         & Spatial    & \vsd{17.33}{38.00} & \vsd{29.33}{58.67} & \vsd{0.00}{4.00} & \vsd{15.55}{33.56} \\
        \hline
        \multirow{5}{*}{Gemini 3.1 Pro}
         & Explicit   & 64.00 & 52.00 & 50.00 & 55.33 \\
         \cline{2-6}
         & Direct     & \vsd{53.33}{10.67} & \vsu{54.67}{2.67} & \vsd{38.67}{11.33} & \vsd{48.89}{6.44} \\
         & Reasoning-Me  & \vsd{56.67}{7.33} & \vsd{51.33}{0.67} & \vsu{55.33}{5.33} & \vsd{54.44}{0.89} \\
         & Contextual & \vsd{58.67}{5.33} & \vsd{50.67}{1.33} & \vsd{44.67}{5.33} & \vsd{51.34}{3.99} \\
         & Spatial    & \vsd{62.67}{1.33} & \vsd{40.67}{11.33} & \vsd{23.33}{26.67} & \vsd{42.22}{13.11} \\
        \hline\hline
        \end{tabular}%
        }
        \end{center}
        \end{table}

\subsection{How Does Longer Context Affect IRG?}

\begin{figure*}[t]
\centering
\includegraphics[width=\textwidth]{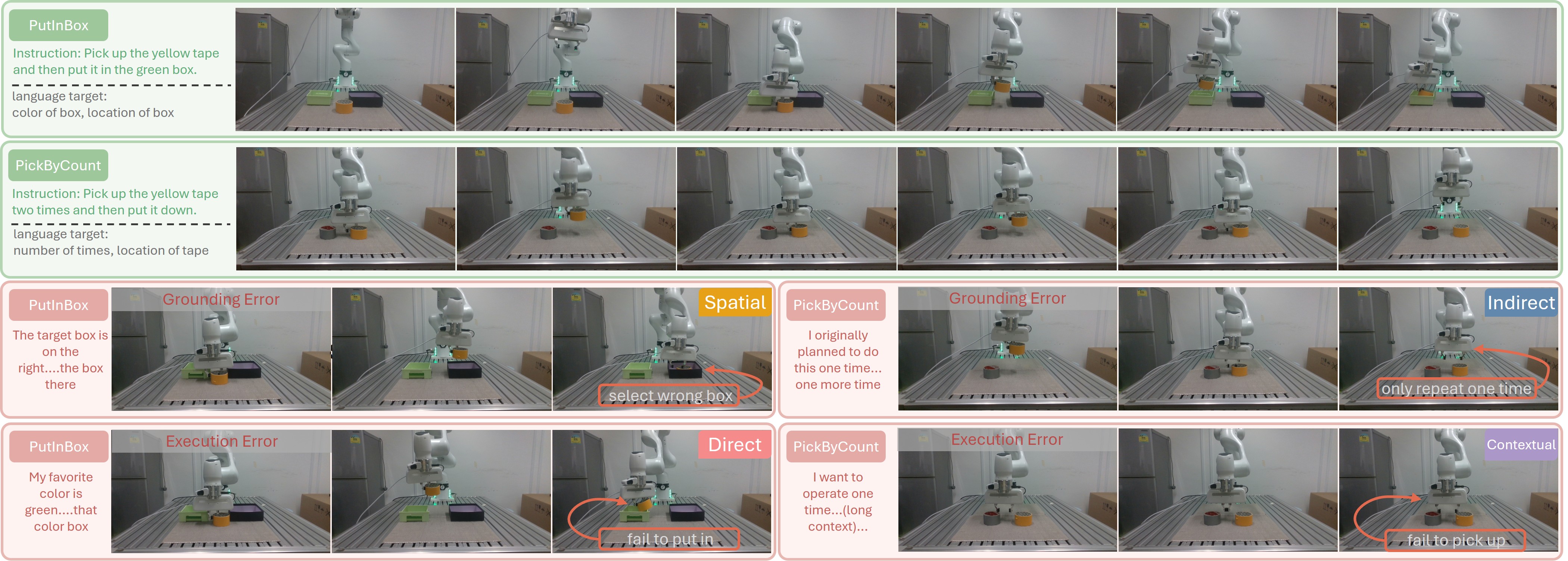}
\caption{Real-world evaluation on PutInBox and PickByCount. Green: representative successful executions under explicit instructions. Red: typical failure cases under IRG, categorized as grounding errors and execution errors.}
\label{fig:real}
\end{figure*}

\begin{table}
        \caption{Effect of increasing linguistic context on Contextual Referential Grounding.
        $\checkmark$~indicates a retained stage: (1) Referent Introduction, (2) Referent Reinforcement, (3) Distractor Context, (4) Action Cue, and (5) Referential Instruction.}
        \label{tab:ctx-ablation}
        \begin{center}
        \footnotesize
        \setlength{\tabcolsep}{2.6pt}
        \setlength{\extrarowheight}{3pt}
        \begin{tabular}{@{}lccccc@{\hspace{5pt}}cccc@{}}
        \hline\hline
         & (1) & (2) & (3) & (4) & (5)
         & \begin{tabular}[c]{@{}c@{}}Bin\\Fill\end{tabular}
         & \begin{tabular}[c]{@{}c@{}}Pick\\Xtimes\end{tabular}
         & \begin{tabular}[c]{@{}c@{}}Swing\\Xtimes\end{tabular}
         & Avg \\
        \hline
        & $\checkmark$ & & & & $\checkmark$ & 24.00 & 15.33 & 48.00 & 29.11 \\
        & $\checkmark$ & $\checkmark$ & & & $\checkmark$ & 30.67 & 15.33 & 44.67 & 30.22 \\
        & $\checkmark$ & $\checkmark$ & $\checkmark$ & & $\checkmark$ & 20.00 & 14.67 & 44.67 & 26.45 \\
        \multirow{-4}{*}{FrameS-Modul} & $\checkmark$ & $\checkmark$ & $\checkmark$ & $\checkmark$ & $\checkmark$ & 21.33 & 13.33 & 38.67 & 24.44 \\
        \hline
        & $\checkmark$ & & & & $\checkmark$ & 41.33 & 81.33 & 8.00 & 43.55 \\
        & $\checkmark$ & $\checkmark$ & & & $\checkmark$ & 44.67 & 62.00 & 8.00 & 38.22 \\
        & $\checkmark$ & $\checkmark$ & $\checkmark$ & & $\checkmark$ & 41.33 & 59.33 & 3.33 & 34.66 \\
        \multirow{-4}{*}{\begin{tabular}[c]{@{}l@{}}GroundSG-\\QwenVL\end{tabular}} & $\checkmark$ & $\checkmark$ & $\checkmark$ & $\checkmark$ & $\checkmark$ & 43.33 & 58.67 & 3.33 & 35.11 \\
        \hline
        & $\checkmark$ & & & & $\checkmark$ & 49.33 & 38.00 & 14.00 & 33.78 \\
        & $\checkmark$ & $\checkmark$ & & & $\checkmark$ & 43.33 & 35.33 & 8.67 & 29.11 \\
        & $\checkmark$ & $\checkmark$ & $\checkmark$ & & $\checkmark$ & 40.67 & 33.33 & 12.67 & 28.89 \\
        \multirow{-4}{*}{MemER} & $\checkmark$ & $\checkmark$ & $\checkmark$ & $\checkmark$ & $\checkmark$ & 34.67 & 37.33 & 5.33 & 25.78 \\
        \hline\hline
        \end{tabular}
        \end{center}
        \end{table}

To examine the effect of extended linguistic context in Contextual Referential Grounding, we progressively insert Referent Reinforcement, Distractor Context, and an Action Cue between the initial referent and the final referential instruction, while keeping the manipulation task unchanged. As shown in Table~\ref{tab:ctx-ablation}, the results exhibit an overall downward trend as more contextual stages are introduced.

\subsection{Real-World Experiments}

We further evaluate whether the referential robustness gap observed in simulation persists in real-world robotic manipulation. Experiments are conducted on a 7-DoF Franka Research 3 robot arm. Demonstrations are collected through GELLO~\cite{gello}, with visual observations captured by a front-view RGBD camera and a wrist-mounted camera at 30 Hz, both of which are Intel RealSense 435IF.

\textbf{Tasks and IRG Construction.} We design two physical manipulation tasks, \textit{PutInBox} and \textit{PickByCount}. In PutInBox, the robot selects between a green and a purple box and places a yellow tape into the target box. In PickByCount, the robot selects between a yellow and a gray tape and performs the corresponding pick-and-place action once or twice. For each task, we construct an Explicit condition and four IRG variants following the benchmark taxonomy. In PutInBox, Reasoning-Mediated grounding identifies the target through color exclusion, while Spatial grounding refers to the left/right location of the boxes. In PickByCount, Reasoning-Mediated grounding derives the required repetition count through numerical reasoning, while Spatial grounding specifies the target tape through its left/right location. Spatial relations are defined in the robot base frame. Importantly, all policies are trained only with explicit instructions, allowing the IRG settings to directly test robustness to previously unseen forms of referential expression.

\textbf{Experimental Protocol.} For each task, we collect 50 demonstrations for fine-tuning. Target colors, spatial arrangements, and execution counts are balanced across training and evaluation. We evaluate two representative policies, $\pi_{0.5}$ and FrameS-Modul. Both models are fine-tuned for 5k steps with a learning rate of $5 \times 10^{-5}$ and a batch size of 48, including linear warmup over the first 300 steps. During real-world evaluation, each instruction variant is tested for 10 trials, with both the action execution horizon and prediction horizon set to 20 steps.

\begin{table}[t]
        \caption{Real-world task success rates (\%) under explicit instructions and four IRG challenges, with each setting evaluated over 10 trials.
        \textcolor{deltaneg}{$\blacktriangledown$}/\textcolor{deltaneu}{\neutri}:
        decrease/no change relative to Explicit.}
        \label{tab:real_world_counts}
        \begin{center}
        \footnotesize
        \begin{tabular}{@{}llccc@{}}
        \hline\hline
         & & PutInBox & PickByCount & Avg \\
        \hline
        \multirow{2}{*}{Explicit}
         & $\pi_{0.5}$ & 90 & 50 & 70 \\
         & FrameS-Modul & 90 & 80 & 85 \\
        \hline
        \multirow{2}{*}{Direct}
         & $\pi_{0.5}$ & \vsn{90}{0} & \vsd{20}{30} & \vsd{55}{15} \\
         & FrameS-Modul & \vsn{90}{0} & \vsd{70}{10} & \vsd{80}{5} \\
        \hline
        \multirow{2}{*}{Reasoning-Me}
         & $\pi_{0.5}$ & \vsd{0}{90} & \vsd{10}{40} & \vsd{5}{65} \\
         & FrameS-Modul & \vsd{0}{90} & \vsd{20}{60} & \vsd{10}{75} \\
        \hline
        \multirow{2}{*}{Contextual}
         & $\pi_{0.5}$ & \vsd{40}{50} & \vsd{20}{30} & \vsd{30}{40} \\
         & FrameS-Modul & \vsd{60}{30} & \vsd{40}{40} & \vsd{50}{35} \\
        \hline
        \multirow{2}{*}{Spatial}
         & $\pi_{0.5}$ & \vsd{30}{60} & \vsd{30}{20} & \vsd{30}{40} \\
         & FrameS-Modul & \vsd{40}{50} & \vsd{70}{10} & \vsd{55}{30} \\
        \hline\hline
        \end{tabular}
        \end{center}
        \end{table}

\textbf{Result and Analysis.} Table~\ref{tab:real_world_counts} shows that the referential robustness gap persists in the physical world. Both policies achieve relatively strong performance under explicit instructions, but success generally decreases when task-relevant information must be recovered implicitly. Reasoning-Mediated referential grounding is particularly challenging. Both models fail completely for PutInBox trials and achieve only 10\% and 20\% success on PickByCount for $\pi_{0.5}$ and FrameS-Modul, respectively. Direct referential grounding is comparatively robust on PutInBox, where both policies retain 90\% success, but remains more challenging on PickByCount. Contextual and spatial referential grounding also lead to clear degradation for both policies.

As illustrated in Figure~\ref{fig:real}, failures can be broadly divided into grounding errors, where the robot selects an incorrect referent, and execution errors, where the intended target is correctly grounded but the subsequent manipulation fails. For example, under Spatial grounding, PutInBox exhibits 69\% wrong-target selections, whereas such errors are not observed in PickByCount, suggesting that referential difficulty can interact strongly with scene and task structure. Meanwhile, execution failures show that even correct referent resolution does not guarantee successful task completion, as errors can still arise during downstream action generation and control.

\section{CONCLUSIONS}

This paper studies implicit referential grounding as an important yet underexplored capability of Vision-Language-Action models. We introduce RoboIRG-Bench, a benchmark that transforms conventional explicit manipulation instructions into implicit referential settings while preserving the underlying task objectives, covering direct, reasoning-mediated, spatial, and contextual grounding challenges. Our experiments reveal a noticeable referential robustness gap: strong performance under explicit instructions does not reliably transfer to settings where task-relevant information must be recovered from linguistic or perceptual context. Although stronger external VLMs improve performance in several cases, significant failures remain, showing that robust IRG cannot be achieved through vision-language scaling alone. Our results further show retaining or retrieving task-relevant memory does not guarantee that a VLA can correctly interpret, ground, and act on an implicit reference to that context.
Real-world evaluations further confirm that such robustness gaps persist in physical manipulation, with failures arising from both incorrect referential grounding and downstream execution. These findings suggest the need for VLA systems that more tightly integrate contextual language understanding, visual grounding, reasoning, and control, and we hope RoboIRG-Bench provides a useful testbed for advancing more natural and reliable human–robot interaction.

\bibliographystyle{IEEEtranBST/IEEEtran}
\bibliography{main}

\end{document}